\PassOptionsToPackage{colorlinks, citecolor = blue, urlcolor = red}{hyperref}
\documentclass[journal]{IEEEtai}
\usepackage{generic}
\usepackage{cite}
\usepackage{amsmath,amssymb,amsfonts}
\usepackage{algorithmic}
\usepackage{graphicx}
\usepackage{algorithm,algorithmic}
\usepackage{orcidlink}
\hypersetup{hidelinks=true}
\usepackage{textcomp}
\let\labelindent\relax
\usepackage{enumitem}
\usepackage{textcomp}
\usepackage{lipsum}
\usepackage{gensymb}
\usepackage{listings}
\usepackage{booktabs}
\usepackage{multirow}
\usepackage{booktabs}
\usepackage{xcolor}
\usepackage{colortbl}  
\usepackage[colorlinks, citecolor = blue, urlcolor = red]{hyperref}

\def\BibTeX{{\rm B\kern-.05em{\sc i\kern-.025em b}\kern-.08em
    T\kern-.1667em\lower.7ex\hbox{E}\kern-.125emX}}

\begin{document}
\title{BrainAid: Efficient Brain Tumor Detection Using Lightweight CNNs and Explainable AI for Resource-Constrained Environments}

\author{Sai Sriram Gonthina\,\orcidlink{0009-0001-6174-8157},
Ross Gore\,\orcidlink{0000-0003-4065-6146},
Bikash Chandra Singh\,\orcidlink{0000-0002-2870-8137},
Pronaya Bhattacharya\,\orcidlink{0000-0002-1206-2298},
Sachin Shetty\,\orcidlink{0000-0002-8789-0610} \IEEEmembership{Senior Member, IEEE},
Abhishek Phadke\,\orcidlink{0000-0001-6420-0499} \IEEEmembership{Member, IEEE},
and Sandip Roy\,\orcidlink{0000-0003-4743-0342}
\thanks{Sai Sriram Gonthina is with the Department of Computer Science and Engineering, IIIT - Naya Raipur, Raipur, Chhattisgarh, India (e-mail: gonthina22100@iiitnr.edu.in).}
\thanks{Ross Gore is with the Center for Secure \& Intelligent Critical Systems, Old Dominion University, Virginia, USA
 (e-mail: rgore@odu.edu).}
\thanks{Bikash Chandra Singh is with the Department of Computer Science, California State University, Fresno, USA
 (e-mail: bisingh@csufresno.edu).}
\thanks{Pronaya Bhattacharya is with the Department of Computer Science and Engineering, Amity School of Engineering and Technology, Amity University, Kolkata, India
(e-mail: pbhattacharya@kol.amity.edu).}
\thanks{Sachin Shetty is with the Center for Secure \& Intelligent Critical Systems, Old Dominion University, Virginia, USA
(e-mail: sshetty@odu.edu).}
\thanks{Abhishek Phadke is with the School of Engineering \& Computing, Christopher Newport University, Virginia, USA
(e-mail: abhishek.phadke@cnu.edu).}
\thanks{Sandip Roy is with the Center for Secure \& Intelligent Critical Systems, Old Dominion University, Virginia, USA
 (e-mail: sroy@odu.edu, sandiproy9500@gmail.com). (Corresponding author: Sandip Roy.)}}

\maketitle

\begin{abstract}
A brain tumor is an unusual growth of cells in the brain that can be cancerous or not, affecting normal brain function and causing serious risk to health. Existing solutions for brain tumor classification, such as expert manual evaluations of magnetic resonance imaging (MRI) scans, are generally inconsistent as different professionals have different criteria and can be resource-intensive, leading to delayed diagnoses. Additionally, while deep learning models have demonstrated potential, they typically require significant computational resources, motivating the need for more efficient and accessible diagnostic tools that can be deployed in real-time healthcare settings. To address these limitations, this study presents BrainAid, which makes efficient use of Convolutional Neural Networks (CNNs) and Vision Transformers (ViTs) for effective classification of four-class classification of glioma, meningioma, pituitary tumor, and no-tumor MRI scans, outperforming all the standard models such as VGG16, Inception-V3, ResNet-50 and so, while integrating Explainable AI (XAI) to improve interpretability. The BrainAid is trained to classify tumors with a high precision of 98\%, achieving an F1 score of 98\% and an accuracy of 98\% using publicly available dataset. The model is deployed on a Raspberry Pi, demonstrating efficiency in low-resource environments. This system shows the feasibility of real-time, edge-based tumor classification, providing quick diagnostic support for medical professionals and enhancing patient care in resource-limited healthcare facilities and underserved areas.
\end{abstract}



\section{Introduction}
\label{sec:introduction}
\IEEEPARstart{A}{Brain} tumor, which can be classified as either benign or malignant, is a case where there is an excessive growth of cells in the brain. This abnormal growth disrupts optimal neurological functioning, leading to serious health issues. Brain tumors can harm healthy tissue, exert pressure on surrounding brain areas, or compress vital regions, ultimately impairing specific functions \cite{ms:NBTS}. 
Brain and other central nervous system (CNS) tumors are the fifth most common cancer. In 2019, there were 347,992 (262,084-388,896) worldwide cases of brain and CNS cancer, a considerable rise (94.35\%) between 1990 to 2019 \cite{b3}. Early detection and identification of the type of brain tumor can increase a person's life expectancy by 1-2 years \cite{article}. According to the \cite{RAGHAVENDRA2023107063}, three imaging methods are frequently used to generate visuals of internal organs: computed tomography (CT), MRI, and positron emission tomography (PET). Among these methods, MRI scans are the most common considering their non-invasiveness and ability to yield high resolution images of the brain region.

Additionally, incorrect identification of a brain tumor can have significant consequences, affecting the patient's probability of survival \cite{b6}. This has led to increased efforts in developing automated image processing technologies to overcome the challenges of manual diagnosis, reducing variability among radiologists. Early approaches use machine learning models to tackle the brain tumor identification task \cite{b7}, \cite{b8}, \cite{b9}, \cite{b10}, \cite{b11}. For example, Manogaran \textit{et al.} \cite{b9} utilized an orthogonal gamma distribution technique to find abnormalities in the MRI brain scans. Kumar \textit{et al.} \cite{b10} proposed a hybrid model where wavelet filters are employed to preprocess MRI scans, followed by feature selection using particle swarm optimization (PSO), and classification using a support vector machine (SVM). Asodekar \textit{et al.} \cite{b11} applied active contour methods, shape feature extraction methods during the feature extraction stage, and fed features into SVM and random forest. Despite their good performance, these machine learning-based methods typically rely significantly on expert-based feature extraction approaches, which are time-consuming and have poor generalization. 

Recently, deep learning techniques have made remarkable progress in the processing of images \cite{b12} and videos \cite{b13}. In \cite{b14}, the authors employ various pre-trained blocks to extract features and convolutional neural network based classifier to classify the brain tumor. Mallick \textit{et al.} \cite{b15} used a deep wavelet autoencoder for feature reduction and a deep neural network for classification task. Furthermore, Raghavan \textit{et al.} \cite{b45} proposed an ensemble method for detecting breast cancer from infrared breast images, and adopted attention-guided Grad-CAM to provide explainability. Wani \textit{et al.} introduced a framework called DeepXplainer \cite{wani2024deepxplainer}, which involves lung cancer detection and SHapley Additive exPlanations (SHAP) analysis for explainations.

\begin{figure}[t!]
  \centering
  \includegraphics[width=0.45\textwidth, height=0.45\textheight, keepaspectratio]{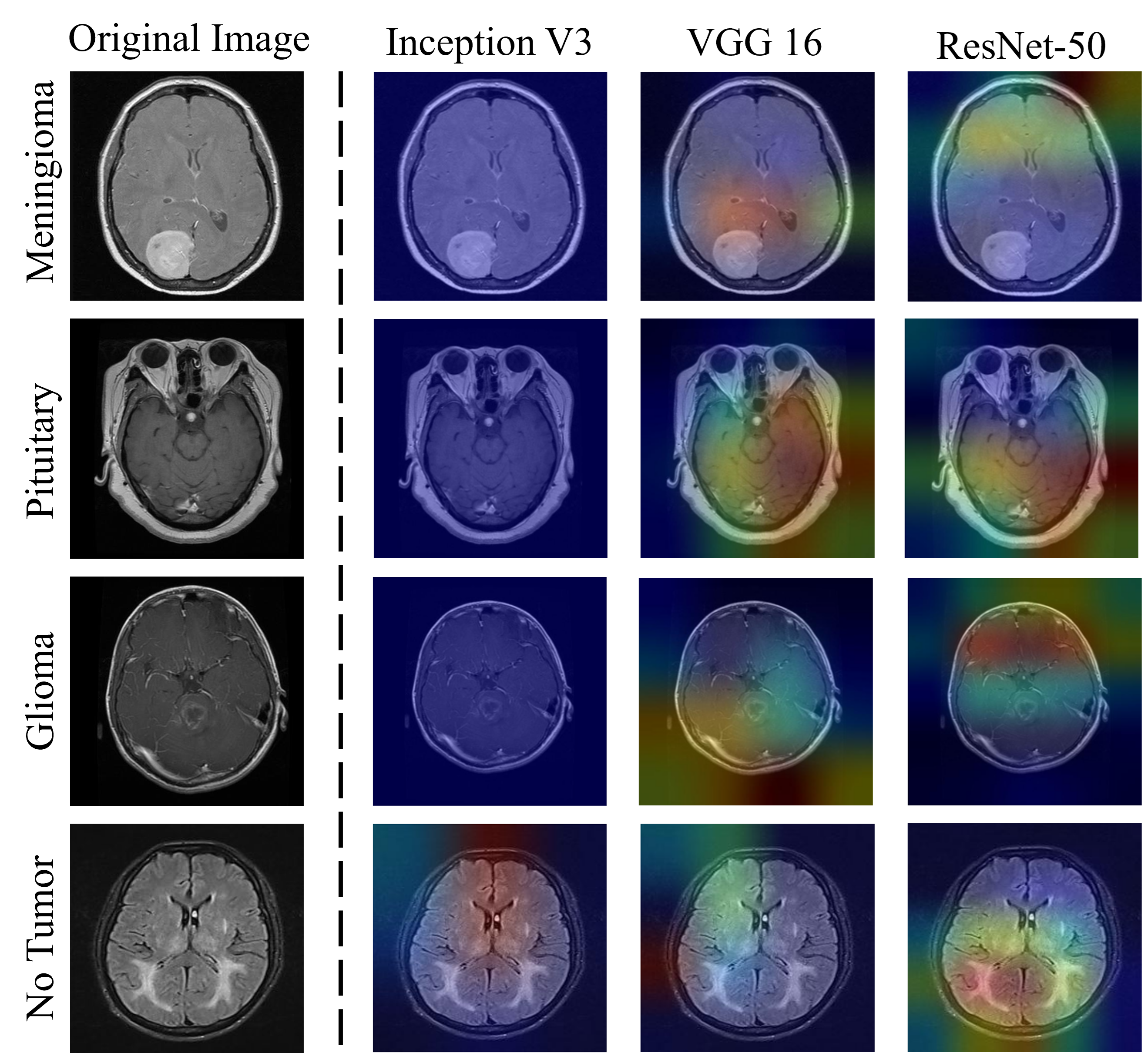}
  \caption{\textbf{Explainability Challenges:} In Grad-CAM visualization, red represents highly activated regions, green represents less activation, and blue represents least focused regions.}
  \label{fig:grad_cam_mot}
\end{figure}

\subsection{Motivation: }

In spite of the good accuracy attained by deep learning (DL) models in identification tasks, they demand more computational resources, making the system computationally expensive. Thus, leading to challenges for implementation in resource-constrained locations like rural hospitals or mobile diagnostic units, where state-of-the-art facilities are absent, making it unsuitable for real-time applications. Therefore, there is a need for developing light-weight models keeping in mind the balance between model complexity and performance. Also, these models often act as black boxes, making it difficult for the user to trust the decision made by the model. As seen in Figure \ref{fig:grad_cam_mot}, there is inconsistency in explainability as the model concentrates on non-tumor areas more than tumor regions, resulting in inaccurate grading. Therefore, this emphasizes the need for transparency and accountability in AI systems. 

Hence, this research focuses on creating a lightweight CNN-based model customized for brain tumor classification that lays a good balance between model complexity and performance. Unlike traditional deep learning techniques, which rely on computationally demanding pre-trained models, the proposed framework is designed for situations that require limited hardware resources. This development solves the limitations of current high-performance DL models by providing a reliable and scalable tool that can be utilized by healthcare professionals and non-specialists alike to accurately detect brain tumors from MRI images, thereby increasing diagnostic reach and consistency in real-world clinical settings.

\subsection{Research Contributions: }
The primary contributions of this study are stated below:
\begin{itemize}

\item \textbf{BrainAid, an end-to-end IoMT framework} that takes a raw MRI scan to a four-class tumor label entirely on the acquisition device, eliminating the need to transmit MRI images to a remote inference server.

\item \textbf{A purpose-built CNN, not a pruned ImageNet backbone.} Filter allocation sized directly for MRI collapses the feature-extractor and classifier stages of existing IoMT pipelines into one 3~million-parameter network.
    
\item \textbf{Explainability analysis beyond classification accuracy.} Cross-model Grad-CAM comparison exposes baselines that reach correct labels while attending to non-tumor tissue, a fail  ure invisible to accuracy alone.

\item \textbf{Testbed validation on resource-constrained edge hardware.} A 13~MB TensorFlow Lite build runs on a Raspberry Pi~4 at 0.05\,s per scan while attaining 98\% accuracy and 98\% F1 across the four classes.

\end{itemize}

The remaining contents of this article will be provided in the following way. Section \ref{sec:Related Works} contains the literature review of prior research for the identification of brain tumors. Section \ref{sec:Methodology} describes the dataset collection, preprocessing, and proposed methodology. Section \ref{sec:Results} discusses the findings and analysis. Section \ref{sec:Deploy} displays the real time deployment and testing of the developed BrainAid. Section \ref{dis} discusses the study's contributions to the clinical field and highlights its limitations. Finally, Section \ref{sec:Conclusion} summarizes the study.

\section{Related Works} \label{sec:Related Works}
Within the domain of brain tumor detection and identification, a variety of methodologies have been explored, each showcasing unique strategies to address the complex challenges faced; the following gives a summary of the merits and shortcomings of each study as shown in Table \ref{Tab_LitSurvey}.

\begin{table*}[!htb]
\caption{A comparative summary of recent schemes in Brain Tumor identification} \label{Tab_LitSurvey}
\centering
\resizebox{0.90\textwidth}{!}{%
\begin{tabular}{| l | c | l | l | l | } \hline
\textbf{Scheme}          &  \textbf{Year}   & \hspace{0.3cm}\textbf{Concept Applied}  & \hspace{0.3cm}\textbf{Major Contributions}              & \hspace{0.3cm} Drawbacks \\  \hline \hline

Anantharajan   &  2024 & * Adaptive Contrast Enhancement      & * Algorithm for noise reduction and contrast  & * Does not incorporate XAI techniques   \\ 
\textit{et al.} \cite{b17}      &  & \hspace{0.2cm}Algorithm   &  \hspace{0.2cm}enhancement of MRI scans      & * Need a significant quantity of training  \\
                    &       & * Gray-Level Co-Occurrence Matrix     &  * Integrates strength of DNN and SVM & \hspace{0.2cm} data to function accurately \cite{b40} \\
                    &       &  * Deep Neural Support Vector Machine             &          &   \\ \hline

Hossain  &  2024     & * Ensemble Learning   & * Included model explainability       & * Increased computational cost due to usage   \\
\textit{et al.} \cite{b19} & & * Transfer Learning       & * Achieved an accuracy of 96.94\%   & \hspace{0.2cm}of multiple models \\
        &           & * Local Interpretable Model-Agnostic   & * Experimentation with multiple CNN    & * Without retraining, the framework land \\
        &           &   \hspace{0.2cm}Explanations (LIME)    &   \hspace{0.2cm}and ViT models          & \hspace{0.2cm} into trouble adjusting to new tumors \cite{b41}\\ \hline

Behera     & 2024  & * Ensemble Transfer Learning   & * Reduced memory overhead imposed by CNN  & * Limited to binary classification problem     \\ 
\textit{et al.}  \cite{b23}  &   & * SLIC superpixel technique   & \hspace{0.2cm} using superpixel algorithm & * Difficult to set optimal hyperparameters  \\
                        &       &                  & * Tested on 3 different datasets    & \hspace{0.2cm} leading to extra overhead\\  \hline

Zhou & 2024   & * Graph Convolution Network       & * Use of spatial and channel-wise convolutions  & * Difficulty segmenting the left edema area   \\ 
\textit{et al.} \cite{b25} &  & * Encoder-decoder framework & \hspace{0.2cm} to extract spatial and temporal data & * Some failed segmentation cases\\
                        &       &                           & * Novel multi-modal correlation loss    & * Complex nature of proposed model \\
                        &       &                           & * Brain Tumor segmentation using graphs    &  \\ \hline

Neamah     & 2024   & * Transfer Learning      & * Overfitting mitigation techniques               & * No interpretability \\
\textit{et al.} \cite{b20}         &        & * Spatial Pyramid Pooling (SPP) & * Diverse parameter analysis for CNN model   & * Limited to CNN framework\\ \hline

Preetha      & 2024  & * Transfer Learning         & * Fine-tuned EfficientNetB4 model       & * Limited to binary classification \\
\textit{et al.}\cite{b21}          &       & * Bayesian optimization          & * Hyperparameter optimization     & * No interpretability \\
                        &       & * Fine tuning               &          &  \\ \hline

Özbay & 2023  & * CNN     & * Classifier with small number of features           & * Unclear performance across diverse \\
 \textit{et al.}  \cite{b24}       &       & * Grad-CAM technique     & * Dimensionality reduction       &  \hspace{0.2cm}datasets\cite{b40} \\
                    &       & * mRMR feature selection method           & * Binary classifier with an accuracy of 99.6\%    & * No type of tumor identification\\ \hline

Sekhar   & 2022 & * Internet of Medical Things (IoMT)  & * GoogleNet used as feature extractor  & * Model tuning and generalization \cite{b42}\\
\textit{et al.} \cite{b22}   &  & * Computer-Aided Diagnostic (CAD)  & * Tested on multiple datasets   & * Lacks XAI model for the reliable  \\ 
                        &       &  \hspace{0.2cm} system           &     & \hspace{0.2cm} identification \cite{b43}\\  \hline

Han \textit{et al.}    & 2019  & * GANs    & * Synthetic dataset generation              & * Artifacts issue like hyper-intensity and  \\
\cite{b44}        &       & * Multimodal unsupervised image-to-   & * GAN-based DA to filter odd  MRI synthetic      & \hspace{0.2cm} gray contours \\
                        &       & \hspace{0.2cm} image translation & \hspace{0.2cm}  images created   & * Increased complexity \\
                                                &       &               &          &  * Manual filtering of training data. \\ \hline
\end{tabular}                                                                                                      
}
\end{table*}

\subsection{Machine Learning-based Diagnosis}
Machine learning (ML) is a subset of artificial intelligence that uses a system to learn and make predictions based on previous instances. 
Mohamed \textit{et al.} \cite{b16} utilizes feature extraction methods like Local Binary Pattern (LBP) and Grey Level Co-occurrence Matrix (GLCM) which then random forest classifier is used to do categorize the MRI scans. However, the system has been developed with a small dataset consisting of only 253 images. The authors in \cite{b17} developed an approach that uses an adaptive contrast enhancement algorithm followed by a combination of neural networks and support vector machine. Though giving accuracy of 97.9\% on the used dataset, these models rely on significant processing resources and a large amount of data to perform optimally. A multi class brain tumor classification system has been proposed by Rolla \textit{et al.} \cite{b18} where the framework achieves an accuracy of 85\% using k-NN classifier. However, extracting relevant features from MRI scans is a difficult task due to its high dimensionality and variability. Minz \textit{et al.} \cite{b32} proposed a binary classifier using GLCM texture feature extractor and Adaboost, achieving an accuracy of 89.9\%. However, the system has been tested with a very limited number of images. Ramasamy \textit{et al.} \cite{b38} used a histogram of oriented gradient (HOG) approach to extract features from MRI images, which were then fed into ML algorithms, with SVM achieving the maximum accuracy. However, the system is limited to binary class classification. In \cite{b33}, the authors presented a performance analysis of 4 different ML models in which SVM attained the highest accuracy of 90\%. However, the study uses a limited number of samples and lacks the diversity in texture properties, which may restrict the model's ability to generalize effectively.

\subsection{Transfer Learning based Diagnosis}
Transfer Learning (TL) is a process in which a model trained on one dataset is fine-tuned for a different task using a target dataset.
Hossain \textit{et al.} \cite{b19} proposes a ensemble of Inception-V3, Xception, VGG16 where the output of these models is fed to a dense layer of a fully connected neural network to give a prediction among 4 classes of the brain tumors. Despite having an F1 score of 75\%, the complexity of this approach adds a lot of overhead to the processing unit due to the size of the learning framework. A binary classification method was proposed by Neamah \cite{b20} where feature extraction of MRI scans is done through a modified ResNet-50 pre-trained model where the final layer is replaced by a spatial pyramid pooling layer to classify the tumors. Nevertheless, the proposed work does not provide any information on the type of brain tumor the patient is suffering from, which is extremely important for diagnosing the patient.  Similarly, Preetha \textit{et al.} \cite{b21} proposed a binary classification system that makes use of the EfficientNet-B4 model and Bayesian optimization to find optimal hyperparameters achieving an accuracy of 99.33\%. Despite attaining high accuracy on one dataset, the work has not been tested with other publicly available datasets to prove its generalizability. 
A pre trained GoogLeNet method was used by the authors in \cite{b22}
to extract features of brain MRI images which was later fed to softmax, SVM, k-NN to classify, from which SVM classifier attained the highest F1 score of 97.35\%. This bimodel setup can complicate the workflow, increasing computational overhead. Asif \textit{et al.} \cite{b26} use DenseNet121, NasNet Large, Xception, and InceptionResNetV2 for feature extraction, in addition to a few dense layers and a softmax layer, to identify tumors as benign or malignant with a 99.67\% accuracy. The key limitation of this work is the small dataset of 1,900 images, which may restrict the model's applicability to larger clinical applications.

\begin{figure*}[h]
\centering
\includegraphics[width=0.95\linewidth]{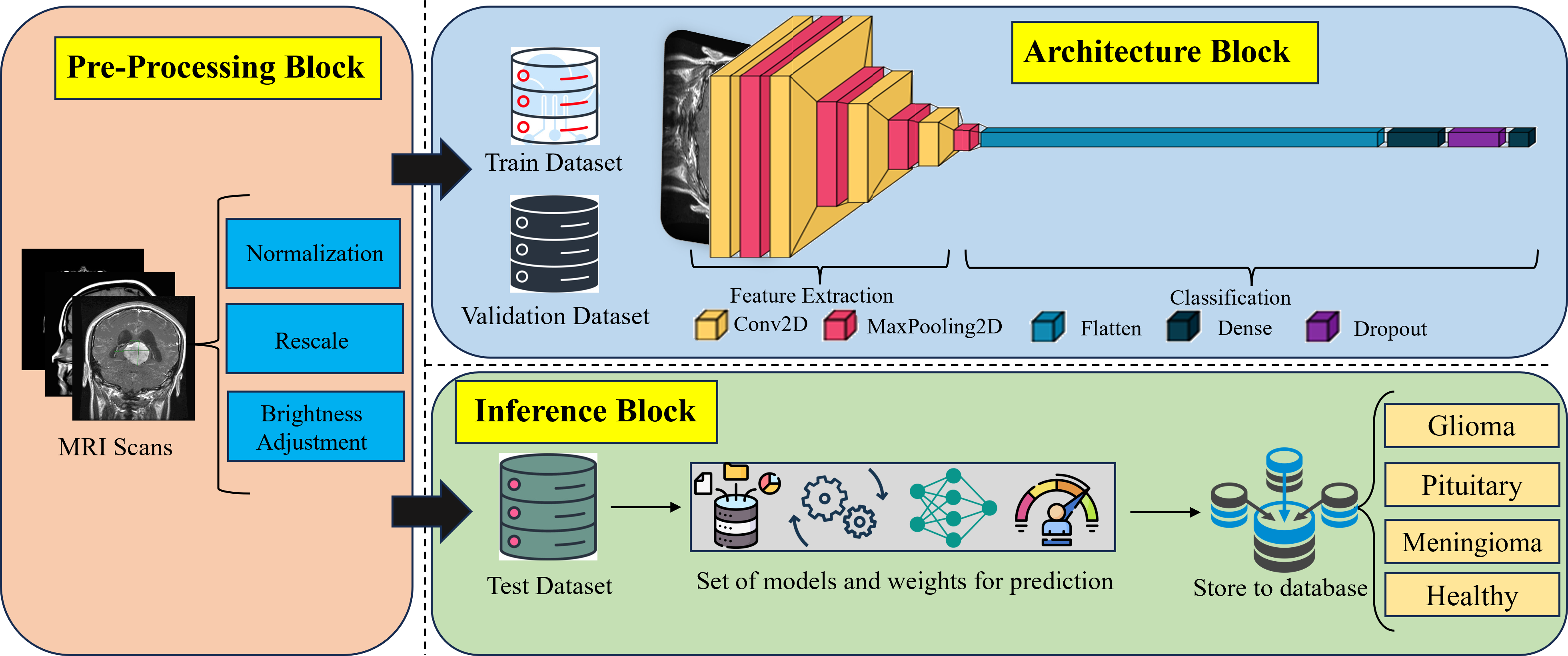}
\setlength{\belowcaptionskip}{-10pt}
\caption{\textbf{Schematic Workflow:} The proposed BrainAid framework, which is primarily composed of three key constituent blocks: (i) the preprocessing block, (ii) the architecture block, and (iii) the interface block.}
\label{fig:flow chart}
\end{figure*}

\subsection{Convolutional Neural Network-based Diagnosis}
A Convolutional Neural Network (CNN) is a deep learning architecture that is especially built to handle structured grid data like videos and images. The approach made by Behera \cite{b23} involves combination of simple linear iterative clustering (SLIC) based superpixel algorithm and CNN for binary classification. However, for super pixelation, choosing the best parameters is a challenging task as it is image-dependent.  Zhou \textit{et al.} \cite{b25} proposed a multi-model graph convolution network to capture spatial and contextual information for tumor segmentation across multiple MRI sequences. However, this method struggles to accurately identify small and scattered tumor regions, which can lead to incomplete segmentation results. Ranjbarzade \textit{et al.} \cite{b27} proposed a classification approach that involves segmenting images into smaller pieces for model training, lowering computation time with the use of a Distance Wise Attention (DWA) algorithm. However, the approach has limits when dealing with tumor volumes that exceed one-third of the brain. In \cite{b28}, an encoder-decoder architecture with three layers was used to capture a variety of characteristics, integrating Dice and focused loss functions. An RF regression model trained on form, volume, and age data obtained 56.7\% training accuracy and 51.7\% validation accuracy. Kobayashi \textit{et al.} \cite{b29} proposed a discriminatory network for detecting regions of tumor by estimating the L2 distance amongst tumor image regions and reconstructed brain image regions in the feature domain.

\textbf{Synthesis: }Numerous brain tumor classification methods have been suggested in the literature so far. Unfortunately, these systems run on models with a high number of parameters and additional feature extraction module, which has an unfavourable impact on the system's overall latency and resource utilisation. As contrast to this, our proposed solution makes use of a custom CNN model with fewer parameters and testbed implementation in the Raspberry Pi to prove its lightweightness, which is essential in IoMT. This makes the system lightweight, autonomous, cost-effective, and real-time.

\section{BrainAid: The proposed sensing system}
\label{sec:Methodology}
We propose BrainAid, a framework that allows doctors and radiologists to easily detect the type of brain tumor. The system classifies brain tumors into four categories: gliomas, meningiomas, pituitary, and no tumor. This categorization of malignant tumors enables doctors to foresee future implications and provide appropriate treatment. Predictions are made locally on the device, eliminating the need to send data to a central server. Figure \ref{fig:flow chart} shows a functional illustration of the process. The detailed steps followed are discussed in this section.

\subsection{Dataset for Training}
The dataset for this research \cite{b30} comprises of 7,023 T1-weighted contrast-enhanced human brain MRI scans classified into four categories: glioma, meningioma, pituitary tumor, and no tumor. This dataset contains three particular kinds of brain tumors, facilitating perform classification tasks in automated brain tumor detection algorithms. It has been split into two groups: a training set of 5,712 images and a testing set of 1,311 images. The training set enabled the deep learning model to acquire convoluted features for accurate tumor classification, whilst the testing set delivered an unbiased assessment of the model's effectiveness. The study used a large dataset to create and test a powerful brain tumor classification model with practical applications.

\subsection{Image Preprocessing}
The preprocessing for the image classification task involves several augmentation and normalization steps to enhance the generalizability of the deep learning model. Initially, images are rescaled by a factor of $\frac{1}{255}$, transforming pixel values from the range [0, 255] to [0, 1], which normalizes the input and helps in faster convergence during training. Then, brightness adjustments within the range [0.85, 1.15] and small translations (width and height shifts) were made using a factor of 0.002 of the total image size. Additionally, shear transformations up to  12.5 degrees and horizontal flips are applied to introduce image variability. 
The mathematical representations of the brightness adjustment transformation applied to image \textit{I} are given below.
\begin{equation}
    I' = I \cdot B, \quad B \in [0.85, 1.15]
\end{equation}

where $I'$ is the adjusted image matrix and $B$ is the brightness factor. This augmentation ensures the model becomes robust to slight variations and reduces overfitting by training on an enriched dataset. The validation and test sets are only rescaled without augmentation, providing a stable benchmark for model evaluation.

\subsection{Proposed BrainAid CNN Model}
\begin{figure}[h]
  \centering
  \includegraphics[width=0.4\textwidth, height=0.45\textheight, keepaspectratio]{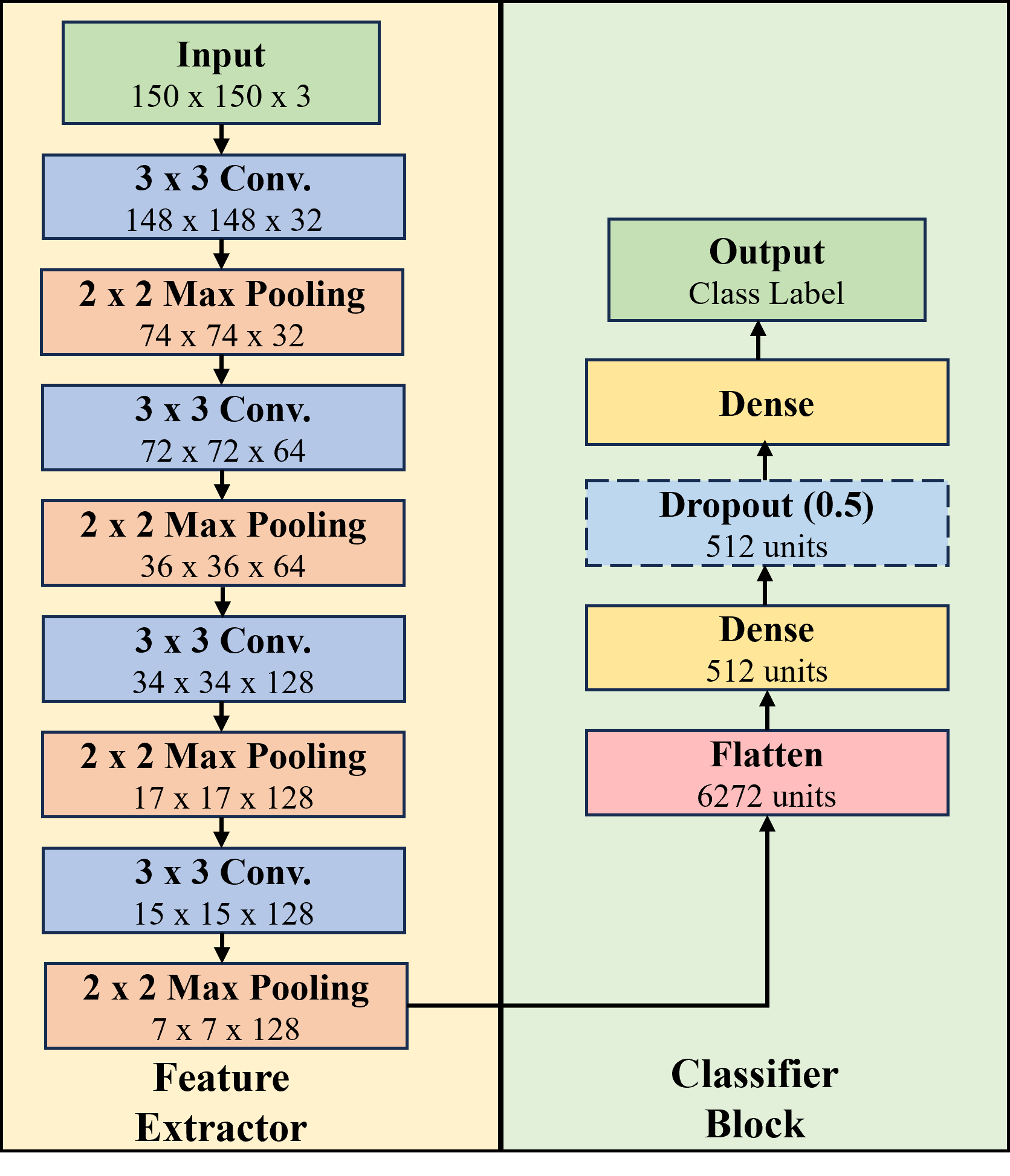}
  \caption{\textbf{BrainAid CNN Architecture: } It is a combination of convolutional 2D, flatten, max pooling, dense, and dropout layers.}
  \label{fig:cnn}
\end{figure}

Artificial intelligence has limited the approach to reproducing a human mind and enabling devices to behave that way. Computer vision and artificial intelligence have been used to solve various crucial challenges. CNNs have been found to be a more effective modeling framework for most applications \cite{b31}. CNNs are algorithms that use deep learning techniques. It accepts an image as input and applies learning weights, allowing the algorithm to distinguish between entities inside a particular frame. We have a picture with RGB information separated into three distinct color planes: red, green, and blue. Figure \ref{fig:cnn} illustrates the offered layout of the network architecture.

The proposed lightweight CNN model uses deep learning techniques to accurately categorize images into groups by finding specific patterns and characteristics. The CNN framework uses convolutional layers to extract significant features from the input image, such as edges, textures, shapes, and gradients. Further, this architecture is optimized through the filter optimization technique, reducing the number of parameters and floating-point operations per second (FLOPs). These traits are compiled by sliding kernels or filters over the input, and the convolution operation is logically characterized as follows:

\begin{equation}
q_{ij}^k = b^k + \sum_{(m, n) \in \mathcal{R}} W_{mn}^k \cdot x_{(i+m)(j+n)},
\end{equation}

The region of interest \(\mathcal{R}\) is defined by the kernel size \(M \times N\). 

The resultant feature maps from this process undergo non-linear activation, typically ReLU (\( f(x) = \max(0, x) \)), which enhances the network's ability to capture complex relationships within the data.

Pooling layers, applied after convolution, further refine the network by reducing spatial dimensions while preserving the most significant features. Max pooling, defined as: 

\begin{equation}
q_{ij} = \max \bigg\{ x_{(2i, 2j)}, x_{(2i+1, 2j)}, x_{(2i, 2j+1)}, x_{(2i+1, 2j+1)} \bigg\},
\end{equation}

minimizes noise and computational cost, enabling efficient edge computing. The flattened feature maps, \( \mathbf{F} = \text{vec}(\mathbf{q}) \), are passed to fully connected Dense layers, where weights \( \mathbf{W} = \{w_{ij} : i, j \in \mathbb{N}\} \) and biases \( \mathbf{b} = \{b_i : i \in \mathbb{N}\} \) optimize feature interdependencies through forward propagation. The inclusion of dropout regularization ensures robustness by reducing overfitting by randomly setting a subset \( \mathcal{D} \subset \{1, 2, \ldots, M\} \) of inputs to zero during training. Finally, the softmax activation function transforms the Dense layer outputs into probabilistic predictions for each class. The backpropagation algorithm fine-tunes weights iteratively to minimize the categorical cross-entropy loss. This pipeline ensures the CNN model is adept at extracting meaningful insights, demonstrating its capability for accurate and efficient image classification. Table \ref{tab:model_settings} shows the parameter values used in BrainAid.

\begin{table}[htbp]
    \centering
    \caption{Description of Training Configuration of the developed BrainAid CNN Model}
    \begin{tabular}{c c c}
        \toprule
        \textbf{Proposed Model} & \textbf{Parameters} & \textbf{Settings} \\
        \midrule
        \multirow{5}{*}{CNN} 
        & Input Size            & $150 \times 150 \times 3$ \\
        & Initial Learning Rate & 0.001 \\
        & Batch Size            & 32    \\
        & Number of Train Batches     & 152    \\
        & Learning Algorithm    & Adam  \\
        & Loss Function         & Cross Entropy \\
        & Best Epoch            & 50    \\
        \bottomrule
    \end{tabular}
    \label{tab:model_settings}
\end{table}

\begin{figure*}[h]
  \centering
  \includegraphics[width=0.75\linewidth]{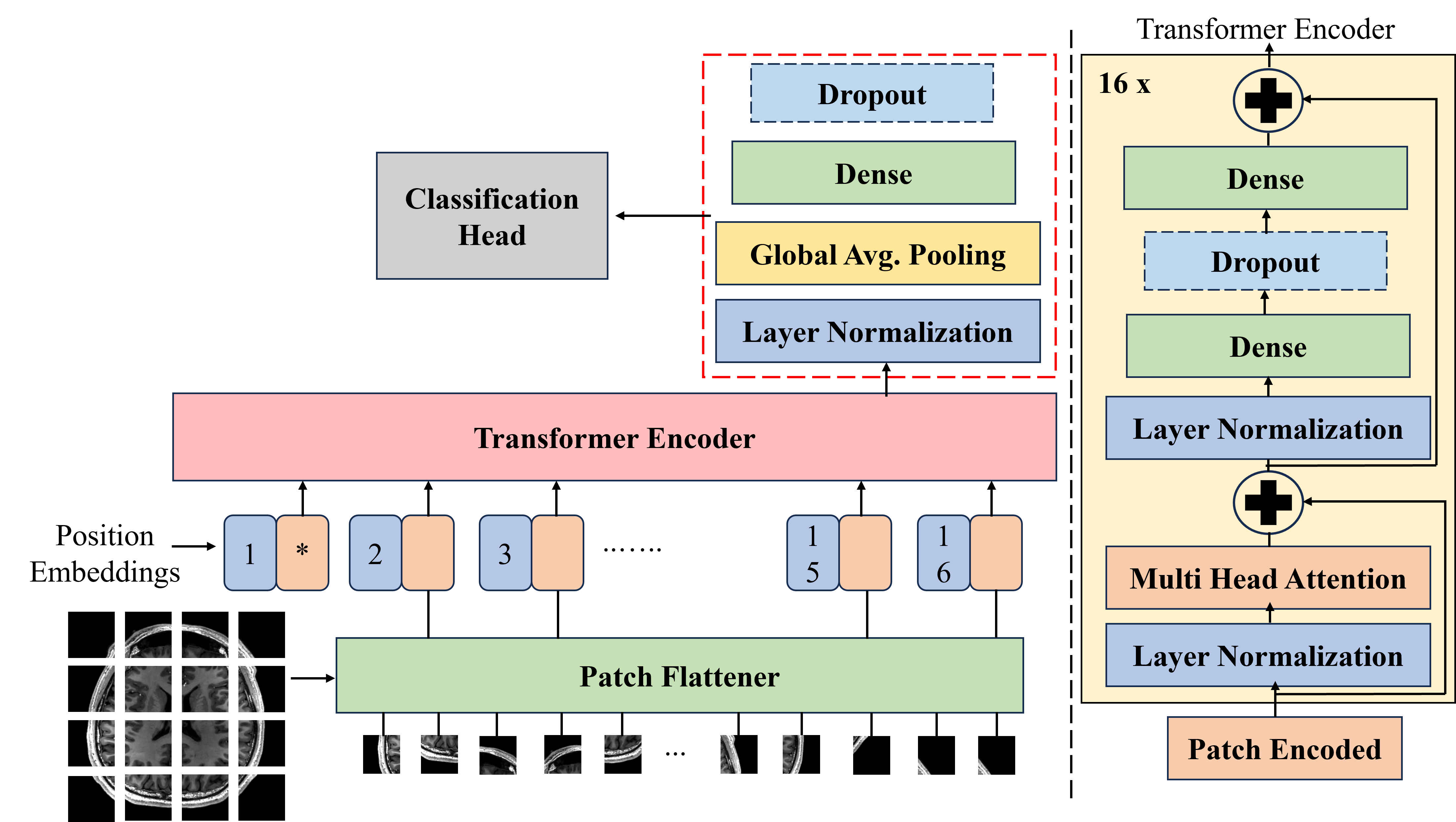}
  \setlength{\belowcaptionskip}{-10pt}
  \caption{\textbf{BrainAid ViT Architecture: } Developed ViT model consists of patch splitting, flattening, embedding followed by transformer encoder for feature extraction task and MLP head for classification of tumors.}
  \label{fig:vit}
\end{figure*}

\subsection{Proposed ViT Model}

The Vision Transformer (ViT) model is a cutting-edge approach to image classification that formulates image data as a sequence of patches. An image \( \textbf{\textit{I}} \in \mathbb{R}^{H \times W \times C} \) is divided into non-overlapping patches of size \( P \times P \), resulting in \( N = \frac{H \cdot W}{P^2} \) patches. Each patch is flattened into a vector \( \mathbf{p}_i \in \mathbb{R}^{P^2 \cdot C} \), and these vectors are projected into a high-dimensional embedding space via a linear layer: 

\begin{equation}
\mathbf{z}_i = \mathbf{W}_E \mathbf{p}_i + \mathbf{b}_E,
\end{equation}

where \( \mathbf{W}_E \in \mathbb{R}^{(P^2 \cdot C) \times D} \) is the learnable weight matrix, and \( \mathbf{b}_E \) is the bias. A positional embedding \( \mathbf{e}_i \in \mathbb{R}^D \) is added to each \( \mathbf{z}_i \) to encode spatial information. The sequence of patch embeddings \( \mathbf{X} = [\mathbf{x}_1, \mathbf{x}_2, \ldots, \mathbf{x}_N] \) serves as the input to the transformer. Within each transformer block, the model applies multi-head self-attention (MHSA), where the attention scores are computed using the query \( \mathbf{Q} \), key \( \mathbf{K} \), and value \( \mathbf{V} \) matrices, as follows:

\begin{equation}
\text{Attention}(\mathbf{Q}, \mathbf{K}, \mathbf{V}) = \text{softmax}\left(\frac{\mathbf{QK}^T}{\sqrt{D}}\right) \mathbf{V},
\end{equation}

where \( \mathbf{Q} = \mathbf{X} \mathbf{W}_Q \), \( \mathbf{K} = \mathbf{X} \mathbf{W}_K \), and \( \mathbf{V} = \mathbf{X} \mathbf{W}_V \), with \( \mathbf{W}_Q, \mathbf{W}_K, \mathbf{W}_V \in \mathbb{R}^{D \times D_h} \) being learnable projection matrices for each head.

The attention output is passed through a feed-forward network (FFN) for feature refinement

\begin{equation}
\text{FFN}(\mathbf{x}) = \text{GELU}(\mathbf{x} \mathbf{W}_1 + \mathbf{b}_1) \mathbf{W}_2 + \mathbf{b}_2,
\end{equation}

where Gaussian Error Linear Unit (GELU) is an activation function, \( \mathbf{W}_1 \in \mathbb{R}^{D \times D_{\text{MLP}}} \), and \( \mathbf{W}_2 \in \mathbb{R}^{D_{\text{MLP}} \times D} \) are learnable weights. To reduce dimensionality, global average pooling aggregates information from all patches. The resulting feature vector is passed through a dense classification head with softmax activation. This mathematical formulation underpins the ViT's capability to achieve state-of-the-art performance on complex image classification tasks. An illustration of the proposed layout of the ViT network is provided in Figure \ref{fig:vit}.

\begin{table*}[!htb]
\caption{Quantitative Results of Compared Methods in Terms of precision , recall, F1 score , Classwise Accuracy, corresponding Weighted Accuracy, and Area Under Curve, in which the Best ones are highlighted in \textcolor{blue}{BLUE}}
\label{tab:model_comparison_cnn}
\centering
\resizebox{0.90\textwidth}{!}{%
\begin{tabular}{|l||l|c|c|c|c|c|c|c|}
\toprule
\textbf{Model} & \textbf{Metric} & \textbf{Glioma} & \textbf{Meningioma} & \textbf{Pituitary} & \textbf{No Tumor} & \textbf{Macro Avg.} & \textbf{Weighted Avg.} & \textbf{AUC} \\
\midrule
\hline
\multirow{3}{*}{ResNet-50} 
& Precision & \cellcolor{cyan!25}\textcolor{blue}{1.00} & 0.91 & 0.91 & 0.99 & 0.95 & 0.95 & \multirow{3}{*}{\textcolor{blue}{0.998}}\\
& Recall    & 0.94 & \cellcolor{cyan!25}\textcolor{blue}{1.00} & \cellcolor{cyan!25}\textcolor{blue}{1.00} & 0.82 & 0.94 & 0.94 & \\
& F1 Score  & 0.97 & 0.95 & 0.95 & 0.89 & 0.94 & 0.94 & \\
\hline
\multirow{3}{*}{VGG16} 
& Precision & 0.85 & 0.87 & 0.84 & 0.88 & 0.86 & 0.86 & \multirow{3}{*}{0.974} \\
& Recall    & 0.98 & 0.98 & 0.88 & 0.55 & 0.85 & 0.86 & \\
& F1 Score  & 0.91 & 0.92 & 0.86 & 0.67 & 0.84 & 0.85 & \\
\hline
\multirow{3}{*}{Inception-V3} 
& Precision & 0.32 & 0.47 & 0.75 & \cellcolor{cyan!25}\textcolor{blue}{1.00} & 0.63 & 0.62 & \multirow{3}{*}{0.803} \\
& Recall    & 0.34 & \cellcolor{cyan!25}\textcolor{blue}{0.99} & 0.34 & 0.00 & 0.42 & 0.46 & \\
& F1 Score  & 0.33 & 0.63 & 0.46 & 0.01 & 0.36 & 0.38 & \\
\hline
\multirow{3}{*}{EfficientNet-B0} 
& Precision & 0.87 & \cellcolor{cyan!25}\textcolor{blue}{0.99} & 0.95 & 0.97 & 0.94 & 0.95 & \multirow{3}{*}{\textcolor{blue}{0.998}} \\
& Recall    & \cellcolor{cyan!25}\textcolor{blue}{1.00} & 0.96 & 0.95 & 0.86 & 0.94 & 0.94 & \\
& F1 Score  & 0.93 & 0.97 & 0.95 & 0.92 & 0.94 & 0.94 & \\
\hline
\multirow{3}{*}{MobileNetV2} 
& Precision & 0.85 & 0.87 & \cellcolor{cyan!25}\textcolor{blue}{1.00} & 0.49 & 0.80 & 0.81 & \multirow{3}{*}{0.957} \\
& Recall    & 0.96 & 0.62 & 0.34 & 0.93 & 0.72 & 0.71 & \\
& F1 Score  & 0.90 & 0.73 & 0.51 & 0.64 & 0.70 & 0.70 & \\
\hline
\multirow{3}{*}{IVX16} 
& Precision & 0.78 & 0.73 & \cellcolor{cyan!25}\textcolor{blue}{1.00} & 0.60 & 0.78 & 0.80 & \multirow{3}{*}{N/A} \\
& Recall    & 0.49 & 0.93 & 0.73 & \cellcolor{cyan!25}\textcolor{blue}{1.00} & 0.79 & 0.76 & \\
& F1 Score  & 0.61 & 0.82 & 0.85 & 0.75 & 0.76 & 0.75 & \\
\hline
\hline
\multirow{3}{*}{\textbf{BrainAid}} 
& Precision & 0.96 & 0.97 & 0.99 & 0.97 & \cellcolor{cyan!25}\textcolor{blue}{0.98} & \cellcolor{cyan!25}\textcolor{blue}{0.98} & \multirow{3}{*}{0.997} \\
& Recall    & \cellcolor{cyan!25}\textcolor{blue}{0.99} & 0.99 & 0.97 & 0.96 & \cellcolor{cyan!25}\textcolor{blue}{0.98} & \cellcolor{cyan!25}\textcolor{blue}{0.98} & \\
& F1 Score  & \cellcolor{cyan!25}\textcolor{blue}{0.98} & \cellcolor{cyan!25}\textcolor{blue}{0.98} & \cellcolor{cyan!25}\textcolor{blue}{0.98} & \cellcolor{cyan!25}\textcolor{blue}{0.96} & \cellcolor{cyan!25}\textcolor{blue}{0.98} & \cellcolor{cyan!25}\textcolor{blue}{0.98} & \\
\bottomrule
\end{tabular}%
}
\end{table*}

\section{Experimental Results and Analysis} \label{sec:Results}

\subsection{Experimental Setup}
The model was trained on a Lenovo LOQ CPU with an Intel Core i7 HX, 16 GB RAM, a 500 GB SSD, and GPU acceleration offered by Kaggle's NVIDIA T4 GPUs in a dual configuration (T4 x2), aggregating 29GB. Python frameworks include Keras, TensorFlow, and TensorFlow Lite. The chipsets utilized to test the BrainAid prototype are the 64-bit quad-core Cortex-A72 processors that exist in the Raspberry Pi 4.

\subsection{Performance and Error Analysis of BrainAid}

\begin{figure}[h]
  \centering
  \includegraphics[width=0.45\textwidth, height=0.45\textheight, keepaspectratio]{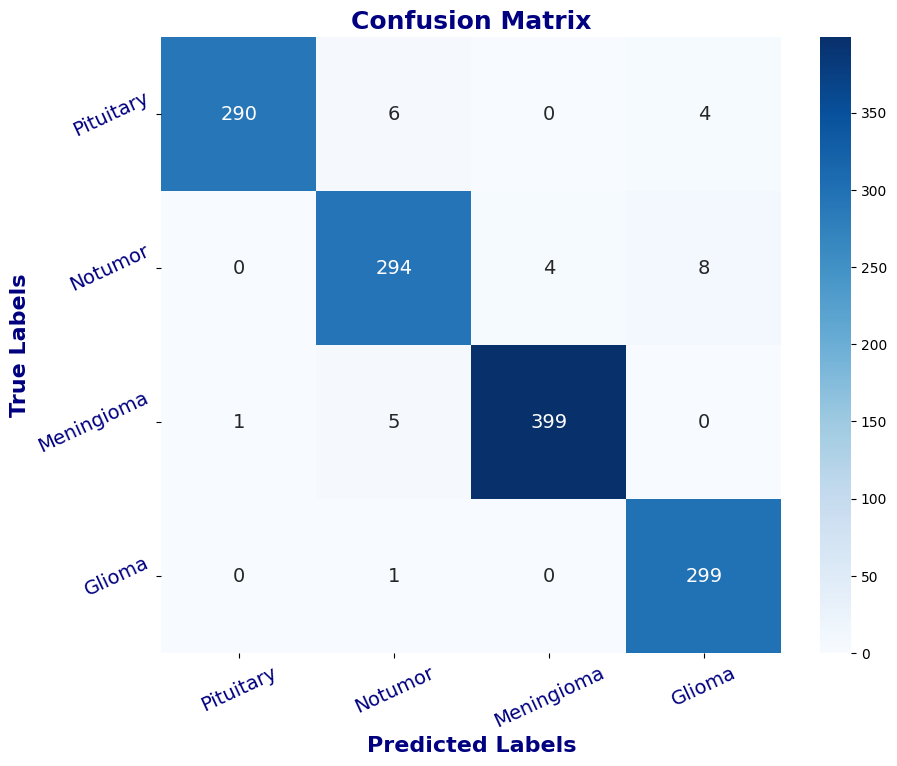}
  \caption{\textbf{Confusion Matrix: } A summary of predictions by the developed BrainAid system.}
  \label{fig:conf_mat}
\end{figure}

\begin{figure}[h]
  \centering
  \includegraphics[width=0.45\textwidth, height=0.4\textheight, keepaspectratio]{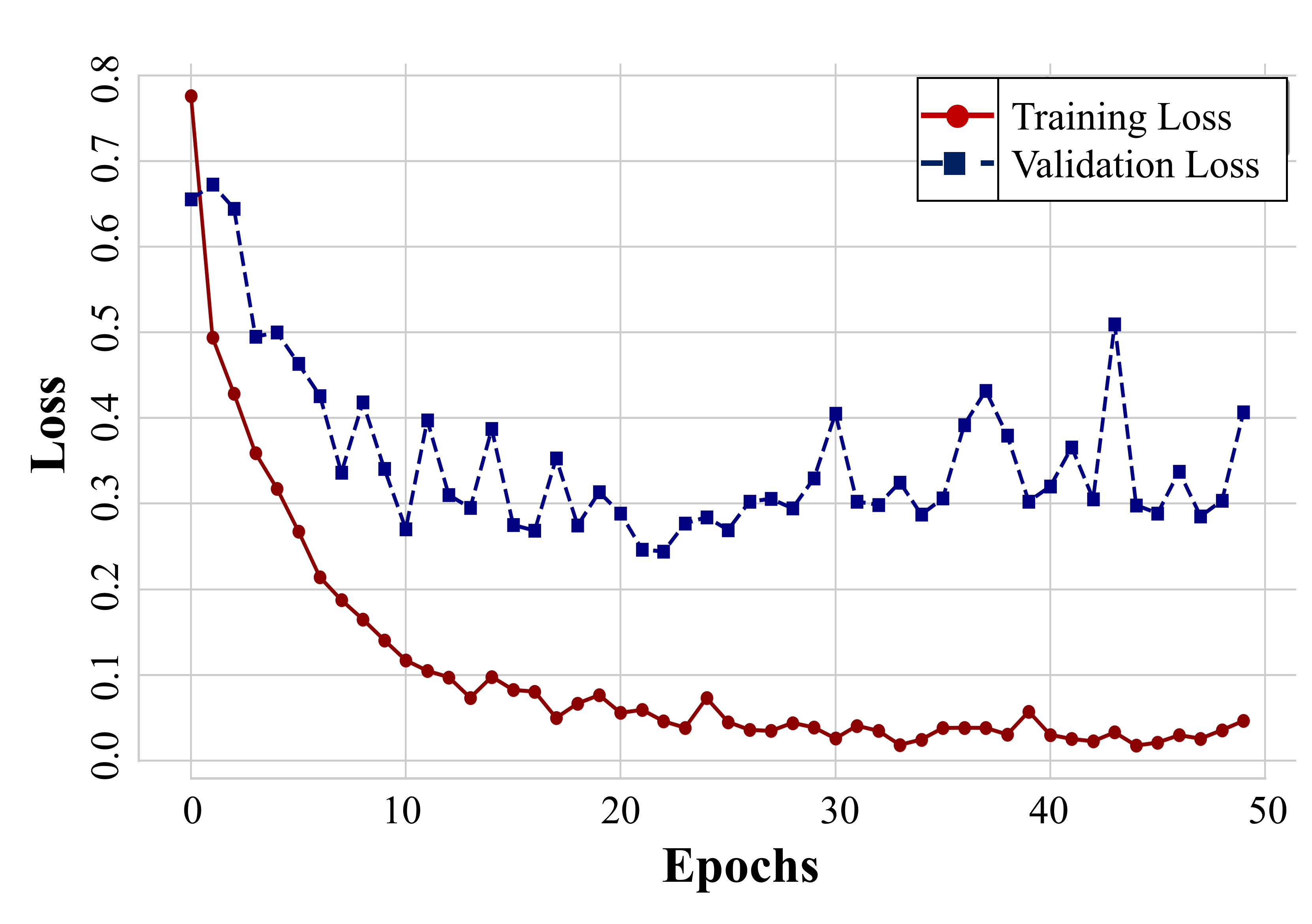}
  \caption{\textbf{Model Development: }The behavior of the BrainAid in loss over the range of 50 epochs.}
  \label{fig:acc}
\end{figure}

\begin{figure}[h]
  \centering
  \includegraphics[width=0.45\textwidth, height=0.4\textheight, keepaspectratio]{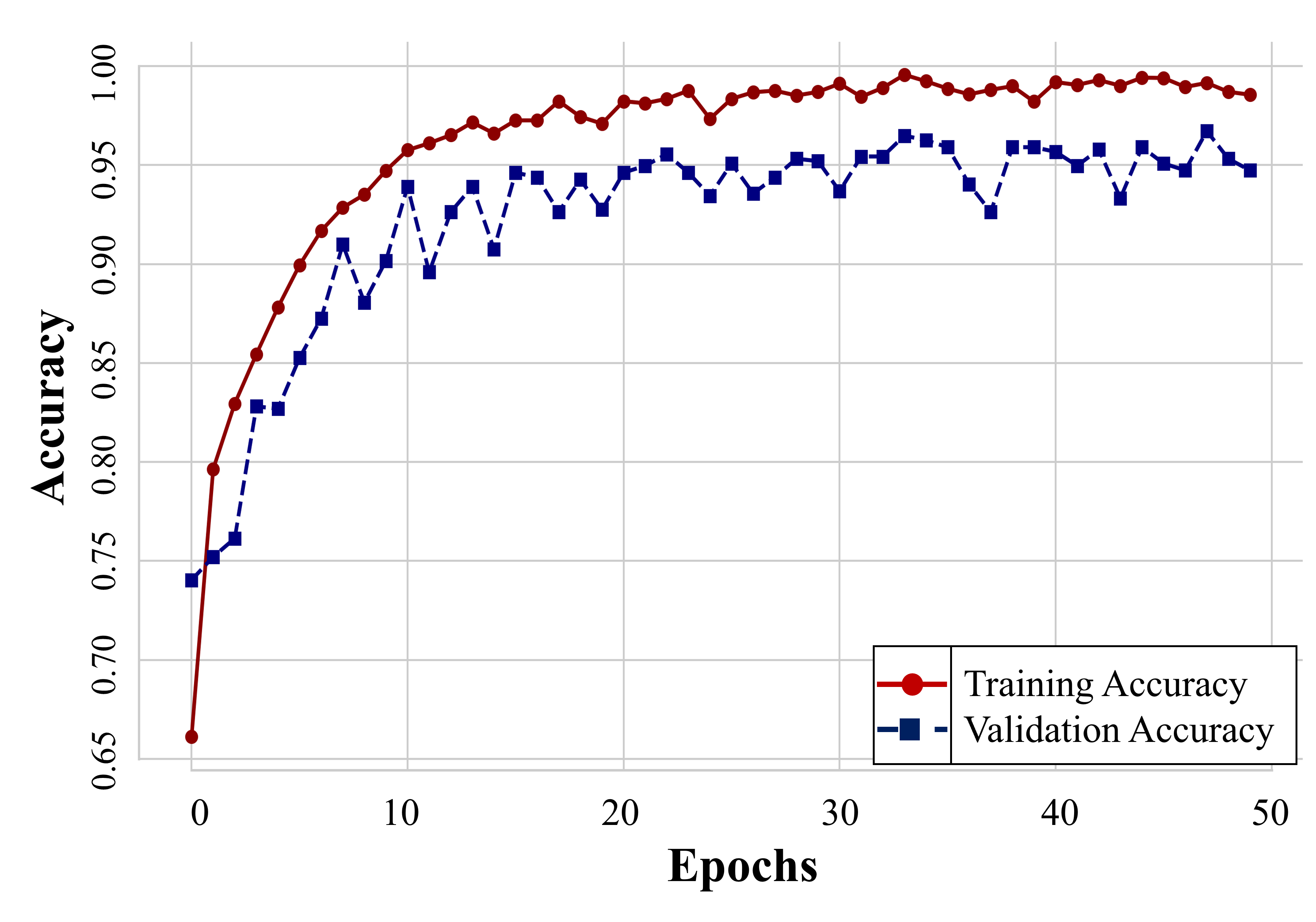}
  \caption{\textbf{Model Development: }The behavior of the BrainAid in accuracy over the range of 50 epochs.}
  \label{fig:acc1}
\end{figure}

The proposed deep learning framework BrainAid underwent through training and evaluation on a diverse set of sample images, representing all target classes. The final results of the CNN BrainAid model and ViT BrainAid model are illustrated in Table \ref{tab:model_comparison_cnn} and Table \ref{tab:model_comparison_vit}, respectively. When benchmarked against existing architectures, the BrainAid model exhibited superior performance, with the next best model, ResNet-50, achieving an accuracy of 94\%. Consequently, while multiple models were assessed for comparative purposes, BrainAid was selected for final deployment due to its high performance and lightweight in nature. Figure \ref{fig:conf_mat} shows the class-wise misclassification across all four labels. 
The training process involved experimenting with various combinations of hyperparameters. Key performance indicators such as training and validation accuracy, training and validation loss, and test accuracy-were thoroughly analyzed to determine the optimal configurations. The accuracy and loss curves for various epoch values are shown in the Figures \ref{fig:acc}, \ref{fig:acc1}. As shown in Table \ref{tab:model_comparison_cnn} for each class, the precision and recall values are high indicating minimal false positive predictions and low false negative predictions. 

Table \ref{tab:model_comparison_vit} highlights improvements in model performance compared to existing work, but some limitations remain. For Glioma, high precision indicates few false positives, but low recall suggests a tendency to miss true cases, requiring focus on improving recall to reduce missed diagnoses. For Meningioma, moderate precision implies some false positives, but high recall demonstrates effective detection of true cases. The model is strong at detecting Pituitary cases with high recall but struggles to discriminate No Tumor instances, most likely due to insufficient negative samples. This is in line with the limitations of using ViT on smaller datasets, as they frequently demand large datasets to work efficiently.

\begin{table*}[!htb]
\caption{Quantitative Results of Compared ViT Methods in Terms of Precision, Recall, F1 Score, Classwise Accuracy, and corresponding Weighted Accuracy in which the Best ones are highlighted in \textcolor{blue}{BLUE} and also shaded}
\label{tab:model_comparison_vit}
\centering
\resizebox{0.90 \textwidth}{!}{%
\begin{tabular}{|l||l|c|c|c|c|c|c|}
\toprule
\textbf{Model} & \textbf{Metric} & \textbf{Glioma} & \textbf{Meningioma} & \textbf{Pituitary} & \textbf{No Tumor} & \textbf{Macro Avg.} & \textbf{Weighted Avg.} \\
\midrule
\hline
SWIN \cite{b19} & Precision & 0.80 & 0.40 & 0.65 & 0.28 & 0.53 & 0.56 \\
              & Recall    & 0.13 & 0.23 & 0.67 & 0.92 & 0.49 & 0.43 \\
              & F1 Score  & 0.22 & 0.30 & 0.66 & 0.43 & 0.40 & 0.39 \\
\hline
CCT \cite{b19}& Precision & 0.74 & \cellcolor{cyan!25}\textcolor{blue}{0.76} & 0.78 & 0.26 & 0.63 & 0.68 \\
             & Recall    & 0.15 & 0.60 & 0.43 & \cellcolor{cyan!25}\textcolor{blue}{0.96} & 0.53 & 0.48 \\
             & F1 Score  & 0.25 & 0.67 & 0.56 & 0.41 & 0.47 & 0.48 \\
\hline
EANet \cite{b19}& Precision & 0.56 & 0.23 & 0.61 & 0.18 & 0.39 & 0.42 \\
               & Recall    & 0.05 & 0.12 & 0.49 & 0.70 & 0.34 & 0.29 \\
               & F1 Score  & 0.10 & 0.15 & 0.54 & 0.28 & 0.27 & 0.27 \\
\hline
\hline
\textbf{ViT-BrainAid} & Precision & \cellcolor{cyan!25}\textcolor{blue}{0.94} & 0.65 & \cellcolor{cyan!25}\textcolor{blue}{0.86} & \cellcolor{cyan!25}\textcolor{blue}{0.63} & \cellcolor{cyan!25}\textcolor{blue}{0.77} & \cellcolor{cyan!25}\textcolor{blue}{0.79} \\
                      & Recall    & \cellcolor{cyan!25}\textcolor{blue}{0.55} & \cellcolor{cyan!25}\textcolor{blue}{0.96} & \cellcolor{cyan!25}\textcolor{blue}{0.95} & 0.69 & \cellcolor{cyan!25}\textcolor{blue}{0.78} & \cellcolor{cyan!25}\textcolor{blue}{0.76} \\
                      & F1 Score  & \cellcolor{cyan!25}\textcolor{blue}{0.69} & \cellcolor{cyan!25}\textcolor{blue}{0.77} & \cellcolor{cyan!25}\textcolor{blue}{0.90} & \cellcolor{cyan!25}\textcolor{blue}{0.66} & \cellcolor{cyan!25}\textcolor{blue}{0.76} & \cellcolor{cyan!25}\textcolor{blue}{0.76} \\
\bottomrule
\end{tabular}%
}
\end{table*}

After extensive experimentation, 50 epochs were determined as the optimal number, beyond which improvements in accuracy were negligible. Dataset distribution played a crucial role in achieving effective model training. While a larger training dataset typically enhances performance, it also increases the risk of overfitting a scenario where the model becomes overly specialized to the training data, leading to poor generalization on real-world test samples. To reduce overfitting and ensure generalization, the dataset was divided into training and test sets in an 80:20 ratio. Then, the training data are split into 90:10 ratio for validation. These validation data are used to monitor and reduce overfitting, while the test data was reserved for the final performance evaluation. A detailed comparative analysis of BrainAid against other state-of-the-art architectures, including EfficientNet-B0, VGG16, Inception-V3, MobileNetV2, and ResNet-50, is presented. The findings clearly demonstrate that BrainAid exceeded all existing frameworks in terms of reliability and accuracy.

\subsection{Model Explainability by Grad-CAM}
Gradient-weighted class activation mapping, often known as Grad-CAM, was initially developed by Selvaraju et al \cite{b39}. and is now commonly utilized to improve the interpretability of DL models. This approach determines what portion of an image significantly impacts decision-making. This is done by obtaining gradient information from the last convolutional layer of the model. Grad-CAM creates heatmaps showing the areas the model focuses on during classification. The heatmaps, shown in Figure 8, overlay essential areas of interest on the original input images, highlighting attributes vital for good prediction. As a result, the predictability and attention mechanism of the model are validated.

\begin{figure*}[h!]
\centering
\includegraphics[width=0.80\linewidth, height=0.42\textheight]{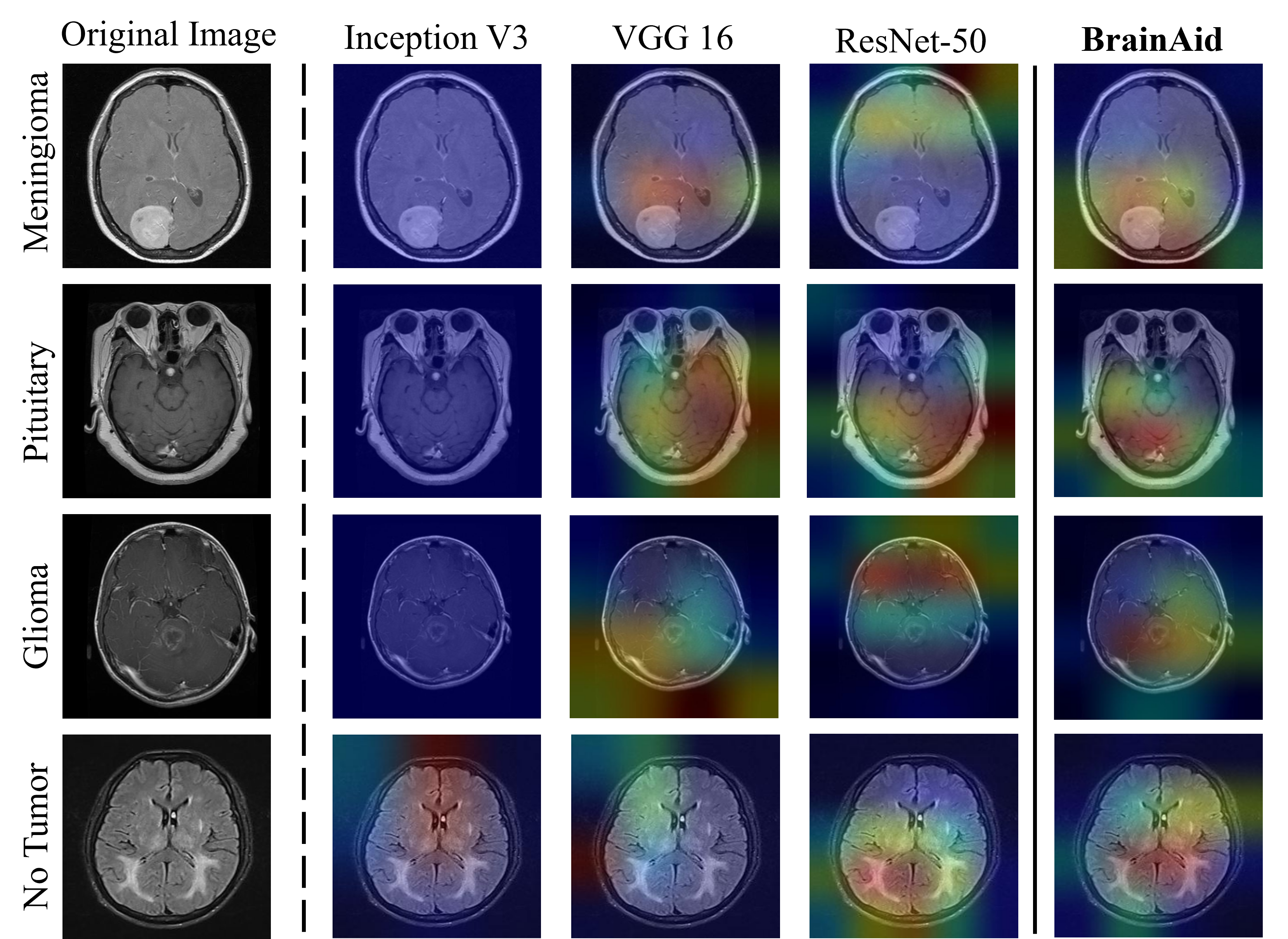}
\setlength{\belowcaptionskip}{-10pt}
\caption{\textbf{Explainability: }this displays the comparison of the key areas targeted by different models, while the developed BrainAid system concentrated on better regions, demonstrating better interpretability.}
\label{fig:grad full}
\end{figure*}

\subsubsection{Meningioma: }
For the meningioma tumor, as we can see in Figure \ref{fig:grad full}, Inception-V3 was unable to focus on important regions as it gave equal importance to whole MRI scan (blue denotes the least focused area). Meanwhile, VGG16 concentrated more on non-tumor regions (red region) than on tumor regions, resulting in poor interpretability. ResNet-50 has 25\% moderately focused (green) regions and 75\% least focused regions. The BrainAid framework created concentrates directly on tumor location and slightly on areas surrounding the tumor, resulting in enhanced explainability.

\subsubsection{Pituitary: }
Next, with regard to of pituitary tumors, Inception-V3 failed to give importance to tumor spots, as confirmed by the blue color of the entire heatmap. In terms of VGG16, the model focuses mostly on the right hemisphere of the brain and moderately on the tumor region. Similarly, ResNet-50 focused mostly on the right side of the tumor and modestly on the tumor region. Unlike prior models, BrainAid focuses the most on tumor and surrounding areas.

\subsubsection{Glioma: }
As we can see, the tumor is located in the center of the brain. However, Inception-V3 fails to correctly detect the tumor in this case as well. VGG16 mistakenly marks the lower left hemisphere region as tumorous and grades modestly on the tumorous portion. Similarly, ResNet-50 highlights more focus on the top region of the brain, where no tumors are found. Finally, BrainAid generates a red patch directly on the tumor as well as some non-tumorous regions, making it a better classifier.

\subsection{Comparison with Prior Art}
The proposed solution BrainAid introduces a lightweight framework to overcome the limitations of traditional brain tumor identification methods, such as the lack of on-device decision-making and reliance on resource-intensive components. By leveraging an Edge-as-a-Service (EaaS) approach, BrainAid provides an end-to-end workflow optimized for edge-based deployment. Its lightweight model design minimizes resource consumption while maintaining high accuracy, which sets it apart from existing techniques. BrainAid achieved an impressive 98\% F1 score in multi-class classification, significantly outperforming methods like IVX16 (75\%)\cite{b19} and the approach by J. Ramasamy \textit{et al.} (87\%)\cite{b38}. The accuracy of each existing technique is shown in Table \ref{tab:method_comparison_acc}. While maintaining superior performance, BrainAid establishes a foundation for developing streamlined, pruned, and downsized deep learning models specifically tailored for edge-based IoMT deployments.

\begin{table}[h]
\caption{Comparison of Brain Tumor Classification Performance With Existing Methods}
\label{tab:method_comparison_acc}
\centering
\setlength{\tabcolsep}{6pt} 
\renewcommand{\arraystretch}{1.1} 
\begin{tabular}{|c||c|c|c|}
\toprule
\textbf{Method} & \textbf{F1 Score} & \textbf{Accuracy} & \textbf{Number of} \\
\textbf{} & \textbf{(\%)} & \textbf{(\%)} & \textbf{Classes} \\
\midrule
\hline
W. Mohamed \textit{et al.} \cite{b16}& 98.3 & 98.0 & 2 \\
K. Neamah \textit{et al.} \cite{b20}& 99.91 & 99.02 & 2 \\
R. Preetha \textit{et al.} \cite{b21}& 99.34 & 99.33 & 2 \\
T. K. Behera \textit{et al.} \cite{b23}& 97.0 & 98 & 2 \\
A. Sekhar \textit{et al.} \cite{b22}& 97.24 & 98.3 & 3 \\
A. Gumaei \textit{et al.} \cite{b46}& 94.5 & 94.23 & 3 \\
A. Younis \textit{et al.} \cite{b47}& 98.31 & 98.32 & 4 \\
S. Hossain \textit{et al.} \cite{b19}& 75.0 & 96.94 & 4 \\
\hline
\hline
\textbf{BrainAid} & \textbf{98.0} & \textbf{98.0} & 4 \\
\bottomrule
\end{tabular}
\end{table}

\section{Prototype Level Deployment of the system}
\label{sec:Deploy}
The BrainAid model created in the TensorFlow framework is converted to TensorFlow Lite format. This process involves freezing the weights and compression of the BrainAid model to optimize it for low computational nodes while ensuring no setbacks in its performance. This extracted model is then deployed in the Raspberry Pi embedded system to demonstrate edge computing and reduce dependency on cloud computing, thus improving privacy concerns. The BrainAid model was compressed to occupy only 13 MB of ROM, enabling efficient storage utilization in IoT systems. Table \ref{tab:model_comparison_param} further compares the BrainAid model to other models in terms of number of parameters, model size, and inference time.

\begin{figure}[h]
    \centering
    \includegraphics[width=8.5cm, height=7cm]{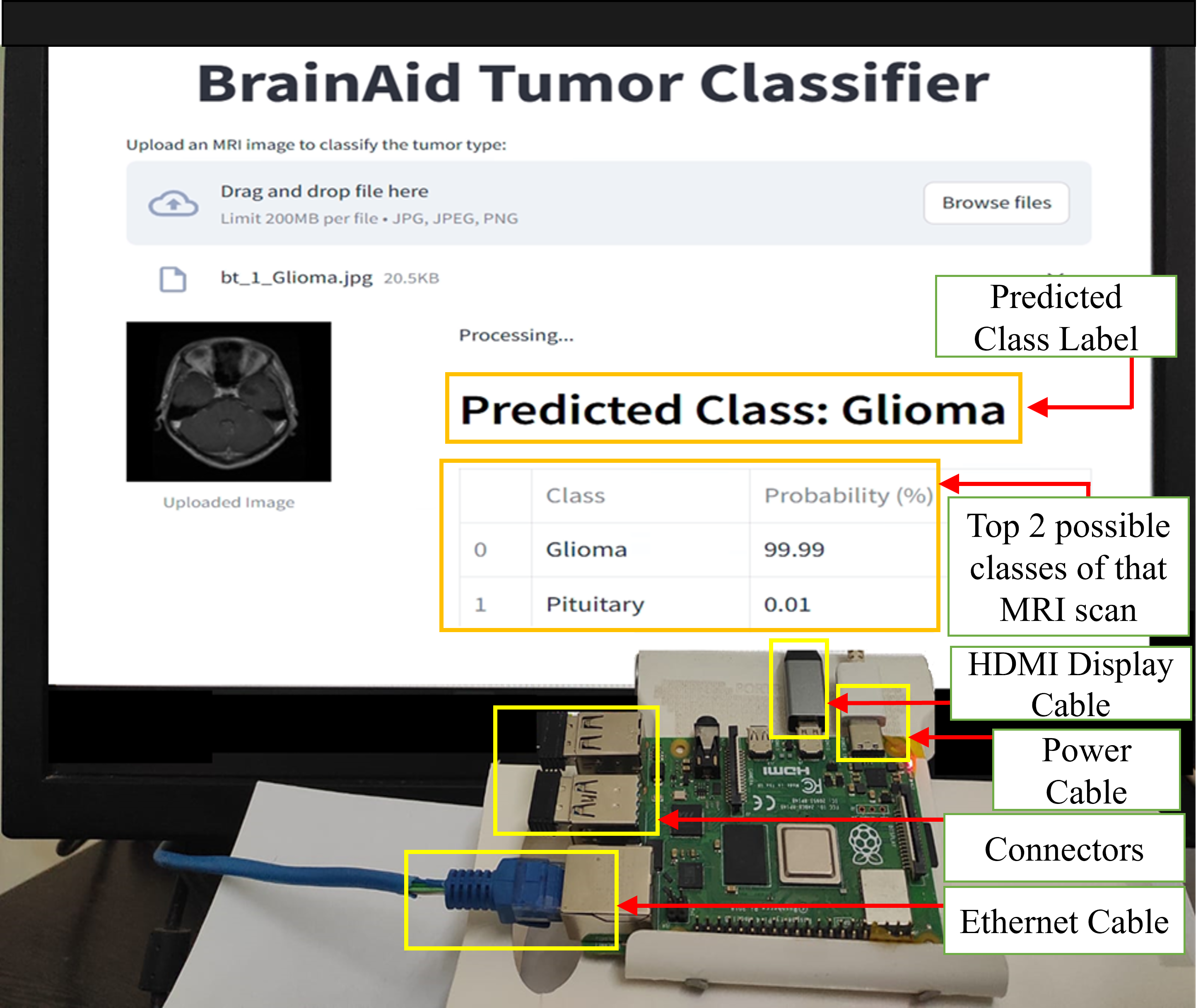}
    \caption{\textbf{Real-time Deployment:} The user selects the MRI scan, the framework then predicts the type of tumor along with the probability score.}
    \label{fig:deploy}
\end{figure}

 The BrainAid system was experimented on a Raspberry Pi 4B model, which features an ARM Cortex-A72 processor with a 1.8 GHz quad-core CPU. The framework was tested under low and high computational resource scenarios to validate its performance and scalability. During testing, BrainAid showed consistent performance with low loss of accuracy, even under resource-constrained settings, highlighting its applicability for edge applications, thus making it ideal for real-world IoT applications. A GUI is created using streamlit to simplify the user experience, and the user first uploads the MRI scan of the brain, then receives the tumor type along with probabilities associated with top two predicted tumor types as shown in Figure \ref{fig:deploy}. Furthermore, deploying the model in an edge-computing environment increases operational autonomy, allowing for better real-time decision-making. This makes BrainAid a great option for resource-constrained areas, allowing for a complete lightweight and efficient diagnostic system for brain tumor categorization.

\begin{table}[htbp]
    \centering
    \caption{Model Comparison on Parameters, Size, and Inference Time}
    \scriptsize
    \begin{tabular}{|l||ccc|}
        \toprule
        \textbf{Model} & \textbf{Params} & \textbf{Model Size} & \textbf{Inference Time} \\
        & \textbf{(Million)} & \textbf{(MB)} & \textbf{(s)} \\
        \midrule
        \hline
        ResNet-50              & 23      & 90  & 3.47 \\
        VGG16                 & 14      & 56  & 0.94 \\
        Inception-V3           & 21.8    & 83  & 5.69 \\
        EfficientNet-B0        & 4       & 15  & 4.93 \\
        MobileNetV2           & 2       & 8   & 0.06 \\
        IVX16 \cite{b19}     & 147     & N/A & $>$6   \\
        GoogLeNet \cite{b22} & 7 & N/A & N/A \\
        \hline
        \hline
        \textbf{BrainAid}                   & 3       & 13  & 0.05    \\
        \bottomrule
    \end{tabular}
    \label{tab:model_comparison_param}
\end{table}

\section{Discussion} \label{dis}

\subsection{Clinical Relevance}
Brain tumor classification is a critical aspect of neuro oncology, as early and accurate diagnosis is important for timely clinical management and treatment planning \cite{b3}. Traditional methods rely on radiologists to interpret MRI scans, which is not only time consuming but also prone to diagnosis variability according to the practitioner's experience. In addition, the global shortage of radiologists is a significant concern, particularly in underserved regions. For example, in some parts of sub-Saharan Africa, there is only one radiologist per million people, highlighting the critical need for accessible diagnostic services in these areas \cite{ms:ramsoft}. Also, in Singapore, there are 7.6 radiologists per 100,000 population, which is lower than Europe's 13 per 100,000 \cite{goh2023growing}. These statistics underscore the pressing need for innovative solutions to address the radiologist shortage and ensure timely medical imaging services worldwide. 

\subsection{Innovation}
The proposed model achieves an impressive F1 score and accuracy of 98\%, achieving competitive classification performance compared with representative deep learning models. By deploying BrainAid on a low-cost Raspberry Pi, we enable real-time, on-site brain tumor detection in environments with scarce high-end computational resources, such as rural clinics and mobile diagnostic units. This demonstrates the potential for deploying computer-assisted MRI analysis in settings with limited computational resources, addressing a critical challenge in underserved areas, namely, the shortage of trained radiologists, which often leads to delays in diagnosis and treatment. Beyond improving accessibility, the model reduces reliance on specialized medical professionals, making it a valuable asset in regions with limited healthcare infrastructure. Additionally, the incorporation of Grad-CAM enhances BrainAid’s transparency by offering visual explanations for its classifications, fostering trust among medical professionals, and assisting them in validating AI-driven decisions.

\subsection{Limitations}
Despite its strong performance, BrainAid has certain limitations. Although ViT have been integrated in low-data system, their performance is suboptimal compared to the CNN model. Most likely due to the relatively small dataset. Larger datasets are needed to fully exploit ViT capabilities. Although the model has been rigorously tested in controlled settings, real world deployment in hospitals and integration with radiology workflows require further validation and regulatory approval. Addressing these limitations will further strengthen the model’s utility and increase its adoption in clinical practice.

\section{Conclusion} \label{sec:Conclusion}
In this study, we explored various SOTA models for brain tumor classification and subsequently proposed a lightweight CNN and ViT architecture, named BrainAid. The proposed model was benchmarked against established algorithms to evaluate its performance. The architecture of BrainAid was designed with lightweight convolutional layers, focusing on achieving an optimal balance between computational efficiency and classification performance. The lightweight nature of the model, combined with strategies such as early stopping, contributed to reducing the risk of overfitting while ensuring robustness against complex image patterns. The results demonstrate that BrainAid outperforms or achieves comparable performance to the aforementioned SOTA algorithms in tumor classification tasks with an accuracy and F1 score of 98\% each. Additionally, we conducted a comparative analysis of the outputs from ViT-based models against those from traditional CNN-based models to assess their relative strengths. To further validate the effectiveness of BrainAid, we employed Grad-CAM, an explainable AI technique, to visualize and analyze the tumor detection capabilities of the model. The Grad-CAM visualizations indicated that BrainAid was more accurate and reliable in identifying tumor regions in medical images compared to the other models. These findings underscore the robustness of the proposed architecture and its potential to contribute to more precise and interpretable brain tumor classification.

\bibliographystyle{IEEEtran}
\bibliography{ref}

\end{document}